\def\year{2021}\relax
\documentclass[letterpaper]{article} % DO NOT CHANGE THIS
\usepackage{aaai21}  % DO NOT CHANGE THIS
\usepackage{times}  % DO NOT CHANGE THIS
\usepackage{helvet} % DO NOT CHANGE THIS
\usepackage{courier}  % DO NOT CHANGE THIS
\usepackage[hyphens]{url}  % DO NOT CHANGE THIS
\usepackage{graphicx} % DO NOT CHANGE THIS
\usepackage{natbib}  % DO NOT CHANGE THIS AND DO NOT ADD ANY OPTIONS TO IT
\usepackage{caption} % DO NOT CHANGE THIS AND DO NOT ADD ANY OPTIONS TO IT
\usepackage{multirow}
\usepackage{subcaption}

\title{CrossNER: Evaluating Cross-Domain Named Entity Recognition}
\author{Zihan Liu, Yan Xu, Tiezheng Yu, Wenliang Dai, Ziwei Ji, \\ Samuel Cahyawijaya, Andrea Madotto, Pascale Fung \\}

\affiliations{
Center for Artificial Intelligence Research (CAiRE)\\
  The Hong Kong University of Science and Technology, Clear Water Bay, Hong Kong\\
  \texttt{zihan.liu@connect.ust.hk, pascale@ece.ust.hk}
 }

\begin{document}

\maketitle

\begin{abstract}
% importance of the cross-domain NER task
Cross-domain named entity recognition (NER) models are able to cope with the scarcity issue of NER samples in target domains.
% problem of current NER datasets
However, most of the existing NER benchmarks lack domain-specialized entity types or do not focus on a certain domain, leading to a less effective cross-domain evaluation.
% Introduce our dataset, CrossNER
To address these obstacles, we introduce a cross-domain NER dataset (CrossNER), a fully-labeled collection of NER data spanning over five diverse domains with specialized entity categories for different domains.
% Introduce domain-related corpus
% In addition, we also provide the domain-related corpus due to the increasing research interests in using it to fine-tune pre-trained language models (\textit{domain-adaptive pre-training}).
Additionally, we also provide a domain-related corpus since using it to continue pre-training language models (\textit{domain-adaptive pre-training}) is effective for the domain adaptation.
% Our experiments & results
We then conduct comprehensive experiments to explore the effectiveness of leveraging different levels of the domain corpus and pre-training strategies to do domain-adaptive pre-training for the cross-domain task.
Results show that focusing on the fractional corpus containing domain-specialized entities and utilizing a more challenging pre-training strategy in domain-adaptive pre-training are beneficial for the NER domain adaptation, and our proposed method can consistently outperform existing cross-domain NER baselines.
Nevertheless, experiments also illustrate the challenge of this cross-domain NER task.
% in the low-resource scenario.
% Nevertheless, the averaged F1-score of our best model is lower than 70\%, which highlights the challenge of this task.
% catalyze research
We hope that our dataset and baselines will catalyze research in the NER domain adaptation area. The code and data are available at \url{https://github.com/zliucr/CrossNER}.

\end{abstract}

\section{Introduction}
% The importance of cross-domain ner
% 1. NER system is important but requires numerous samples to train and hard to generalize to target domains.
% 2. Collecting data is expensive and time-consuming.
Named entity recognition (NER) is a key component in text processing and information extraction. Contemporary NER systems rely on numerous training samples~\cite{ma2016end,lample2016neural,chiu2016named,dong2016character,yadav2018survey}, and a well-trained NER model could fail to generalize to a new domain due to the domain discrepancy. However, collecting large amounts of data samples is expensive and time-consuming. Hence, it is essential to build cross-domain NER models that possess transferability to quickly adapt to a target domain by using only a few training samples.
% we expect NER models trained in the source domain to possess the ability to fast adapt to a target domain by using only a few additional training samples.

% Problems of existing NER datasets w.r.t the cross-domain NER task (target domain that has different categories are limited in biomedical)
Existing cross-domain NER studies~\cite{yang2017transfer,jia2019cross,jia2020multi} consider the CoNLL2003 English NER dataset~\cite{tjong2003introduction} from Reuters News as the source domain, and utilize NER datasets from Twitter~\cite{derczynski2016broad,lu2018visual}, biomedicine~\cite{nedellec2013overview} and CBS SciTech News~\cite{jia2019cross} as target domains. 
% Two issues of current NER datasets
% Although these datasets are available for evaluating the NER domain adaptation, we find out two issues in it.
However, we find two drawbacks in utilizing these datasets for cross-domain NER evaluation.
First, most target domains are either close to the source domain or not narrowed down to a specific topic or domain. Specifically, the CBS SciTech News domain is close to the Reuters News domain (both are related to news) and the content in the Twitter domain is generally broad since diverse topics will be tweeted about and discussed on social media. 
% Hence, more domains that are diverse from the source domain are needed.
Second, the entity categories for the target domains are limited. Except the biomedical domain, which has specialized entities in the biomedical field, the other domains (i.e., Twitter and CBS SciTech News) only have general categories, such as person and location. However, we expect NER models to recognize certain entities related to target domains.

% Introduce our dataset, CrossNER
In this paper, we introduce a new human-annotated cross-domain NER dataset, dubbed CrossNER, which contains five diverse domains, namely, politics, natural science, music, literature and artificial intelligence (AI). And each domain has particular entity categories; for example, there are ``politician'', ``election'' and ``political party'' categories specialized for the politics domain. As in previous works, we consider the CoNLL2003 English NER dataset as the source domain, and five domains in CrossNER as the target domains.
We collect $\sim$1000 development and test examples for each domain and a small size of data samples (100 or 200) in the training set for each domain since we consider a low-resource scenario for target domains.
% cross-domain NER models are anticipated to generalize well to target domains by using a small amount of additional examples.
% Introduce domain-related corpus
In addition, we collect the corresponding five unlabeled domain-related corpora for the domain-adaptive pre-training, given its effectiveness for domain adaptation~\cite{beltagy2019scibert,donahue2019lakhnes,lee2020biobert,gururangan-etal-2020-dont}.
% fine-tuning pre-trained language models~\cite{beltagy2019scibert,lee2020biobert,gururangan-etal-2020-dont}.

% our experiments
We evaluate existing cross-domain NER models on our collected dataset and 
% propose competitive baselines for this task. 
explore using different levels of domain corpus and masking strategies to continue pre-training language models (e.g., BERT~\cite{devlin2019bert}). Results show that emphasizing the partial corpus with specialized entity categories in BERT's domain-adaptive pre-training (DAPT) consistently improves its domain adaptation ability. Additionally, in the DAPT, BERT's masked language modeling (MLM) can be enhanced by intentionally masking contiguous random spans, rather than random tokens.
Comprehensive experiments illustrate that the span-level pre-training consistently outperforms the original MLM pre-training for NER domain adaptation.
% , especially when a large size of domain-related corpus is not available.
% However, the domain adaptation performance is still a long way from satisfactory, which illustrates the challenge of this task and highlights that more advanced models are needed. 
% However, the F1-score of our best cross-domain model (averaged over five domains) is lower than 70\%, which illustrates the challenge of this task and highlights that more advanced models are needed. 
Furthermore, experimental results show that the cross-domain NER task is challenging, especially when only a few data samples in the target domain are available.
% , pointing to avenues for future research in this area.
The main contributions of this paper are summarized as follows:
\begin{itemize}
    \item We introduce CrossNER, a fully-labeled dataset spanning over five diverse domains, as well as the corresponding five domain-related corpora for studies of the cross-domain NER task.
    \item We report a set of benchmark results of existing strong NER models, and propose competitive baselines which outperform the current state-of-the-art model.
    \item To the best of our knowledge, we are the first to conduct in-depth experiments and analyses in terms of the number of target domain training samples, the size of domain-related corpus and different masking strategies in the DAPT for the NER domain adaptation.
\end{itemize}

\section{Related Work}
\subsubsection{Existing NER Datasets}
% 1. list NER datasets
CoNLL2003~\cite{tjong2003introduction} is the most popular NER dataset and is collected from the Reuters News domain. It contains four general entity categories, namely, person, location, organization and miscellaneous. The Email dataset~\cite{lawson2010annotating}, Twitter dataset~\cite{derczynski2016broad,lu2018visual}, and SciTech News dataset~\cite{jia2019cross} have the same or a smaller set of entity categories than CoNLL2003.
WNUT NER~\cite{strauss2016results}, from the Twitter domain, has ten entity types.\footnote{Ten categories: organization, location, person, facility, movie, music artist, product, sports team, TV show and miscellaneous.} However, aside from four entity types that are the same as those in CoNLL2003, the other six types come from four domains, which means they are not concentrated on a particular domain.
Different from these datasets, the OntoNotes NER dataset~\cite{pradhan2012conll} consists of six genres (newswire, broadcast news, broadcast conversation, magazine, telephone conversation and web data). However, the six genres are either relatively close (e.g., newswire and broadcast news) or have broad content (e.g., web data and magazine). 
% Hence, existing NER datasets that are suitable for domain adaptation are limited.
To the best of our knowledge, only Biomedical NER~\cite{nedellec2013overview} and CORD-NER~\cite{wang2020comprehensive} focus on specific domains (biomedicine and COVID-19, respectively) which are diverse from the News domain and have specialized entity classes. However, annotations in CORD-NER are produced by models instead of annotators.

\subsubsection{Cross-Domain NER}
Cross-domain algorithms alleviate the data scarcity issue and boost the models' generalization ability to target domains~\cite{kim2015new,yang2018design,lee2018transfer,lin2018neural,liuzeroresource}.
\citet{daume2007frustratingly} enhanced the adaptation ability by mapping the entity label space between the source and target domains. \citet{wang2018label} proposed a label-aware double transfer learning framework for cross-specialty NER, while \citet{wang2020multi} investigated different domain adaptation settings for the NER task. \citet{liu-etal-2020-coach} introduced a two-stage framework to better capture entities for input sequences.
\citet{sachan2018effective, jia2019cross} injected target domain knowledge into language models for the fast adaptation, and \citet{jia2020multi} presented a multi-cell compositional network for NER domain adaptation.
% \subsubsection{Fast Adaptation Systems}
% Tackling low-resource issues has always been an interesting yet challenging task~\cite{pan2010cross,jaech2016domain}, and 
Additionally, fast adaptation algorithms have been applied to low-resource languages~\cite{lample2019cross,liu2019zero,liu2020attention,liu2020cross,wilie2020indonlu}, accents~\cite{Winata2020}, and machine translation~\cite{artetxe2018unsupervised,lample2018unsupervised}.

\begin{table*}[]
\renewcommand{\arraystretch}{1.15}
\centering
\resizebox{0.968\textwidth}{!}{
\begin{tabular}{|c|c|c|c|c|c|c|c|}
\hline
\multirow{2}{*}{\textbf{Domain}}                                  & \multicolumn{3}{c|}{\textbf{Unlabeled Corpus}} & \multicolumn{3}{c|}{\textbf{Labeled NER}} & \multirow{2}{*}{\textbf{Entity Categories}}    \\ \cline{2-7}
  & \textbf{\# paragraph}    & \textbf{\# sentence}    & \textbf{\# tokens}   & \textbf{\# Train}         & \textbf{\# Dev}         & \textbf{\# Test}        & \\ \hline
Reuters & - & - & - & 14,987 & 3,466 & 3,684 & person, organization, location, miscellaneous \\ \hline
Politics        & 2.76M           & 9.07M           & 176.56M      & 200           & 541         & 651         & \begin{tabular}[c]{@{}c@{}}politician, person, organization, political party, event, \\ election, country, location, miscellaneous\end{tabular}    \\ \hline
\begin{tabular}[c]{@{}c@{}}Natural\\ Science\end{tabular}         & 1.72M           & 5.32M           & 98.50M       & 200           & 450         & 543         & \begin{tabular}[c]{@{}c@{}}scientist, person, university, organization, country, location, discipline, \\ enzyme, protein, chemical compound, chemical element, event,\\  astronomical object, academic journal,  award, theory, miscellaneous\end{tabular} \\ \hline
Music      & 3.49M           & 9.82M           & 194.62M       & 100           & 380         & 456         & \begin{tabular}[c]{@{}c@{}}music genre, song, band, album, musical artist, musical instrument,\\ award, event, country, location, organization, person, miscellaneous\end{tabular}     \\ \hline
Literature    & 2.69M           & 9.17M           & 177.33M       & 100           & 400         & 416         & \begin{tabular}[c]{@{}c@{}}book, writer, award, poem, event, magazine, person, location,\\  organization, country, miscellaneous\end{tabular}    \\ \hline
\begin{tabular}[c]{@{}c@{}}Artificial\\ Intelligence\end{tabular} & 97.04K           & 287.62K           & 5.20M       & 100           & 350         & 431         & \begin{tabular}[c]{@{}c@{}}field, task, product, algorithm, researcher, metrics, university, \\ country, person, organization, location, miscellaneous\end{tabular}     \\ \hline
\end{tabular}
}
\caption{Data statistics of unlabeled domain corpora, labeled NER samples and entity categories for each domain.}
\label{table:datastatistics}
\end{table*}

\section{The CrossNER Dataset}
To collect CrossNER, we first construct five unlabeled domain-specific (politics, natural science, music, literature and AI) corpora from Wikipedia. Then, we extract sentences from these corpora for annotating named entities. The details are given in the following sections.

\subsection{Unlabeled Corpora Collection}
Wikipedia contains various categories and each category has further subcategories. It serves as a valuable source for us to collect a large corpus related to a certain domain. For example, to construct the corpus in the politics domain, we gather Wikipedia pages that are in the politics category as well as its subcategories, such as political organizations and political cultures. 
% More detailed information is reported in Appendix A. 
We utilize these collected corpora to investigate domain-adaptive pre-training.

\subsection{NER Data Collection}
\subsubsection{Pre-Annotation Process}
For each domain, we sample sentences from our collected unlabeled corpus, which are then given named entity labels. Before annotating the sampled sentences, we leverage the DBpedia Ontology~\cite{mendes2012dbpedia}\footnote{The DBpedia Ontology contains 320 entity classes and categorizes 3.64 million entities.} to automatically detect entities and pre-annotate the selected samples. By doing so, we can alleviate the workload of annotators and potentially avoid annotation mistakes. However, the quality of the pre-annotated NER samples is not satisfactory since some entities will be incorrectly labeled and many entities are not in the DBpedia Ontology. In addition, we utilize the hyperlinks in Wikipedia and mark tokens that have hyperlinks to facilitate the annotation process and assist annotators in noticing entities. This is because tokens having hyperlinks are highly likely to be the named entity.
% Hence, we mark tokens that have hyperlinks before starting annotations.
\subsubsection{Annotation Process}
Each data sample requires two well-trained NER annotators to annotate it and one NER expert to double check it and give final labels. The data collection proceeds in three steps. First, one annotator needs to detect and categorize the entities in the given sentences. Second, the other annotator checks the annotations made by the first annotator, makes markings if he/she thinks that the annotations \textbf{could} be wrong and gives another annotation. Finally, the expert first goes through the annotations again and checks for possible mistakes, and then makes the final decision for disagreements between the first two annotators. 
% Our data collection steps are modified based on~\citet{schuster2019cross}. 
In order to ensure the quality of annotations, the second annotator concentrates on looking for possible mistakes made by the first annotator instead of labeling from scratch. In addition, the expert will give a second round check and confer with the first two annotators when he/she is unsure about the annotations. A total 63.64\% of entities (the number of pre-annotated entities divided by the number of entities with hyperlinks) are pre-annotated based on the DBpedia Ontology, 73.33\% of entities (the number of corrected entities divided by the number of entities with hyperlinks) are corrected in the first annotation stage, 8.59\% of entities are annotated as possibly incorrect in the second checking stage, and finally, 8.57\% of annotations (out of all annotations) are modified by the experts. The details are reported in the Appendix.
% The detailed information about collecting CrossNER is in the Appendix.
% The detailed annotation instruction is in the Appendix B.

\subsection{Data Statistics}
% 1. Number of data samples for each domain (checked)
% 2. Entity types for each domain (checked) => talk how we select entity categories?
% 3. Number of sentences for unlabeled corpus (checked)
% 4. data statistics for each entity type (maybe appendix)
% 5. consider to put NER examples for each domain

% Need to discuss
% 1. unlabeled corpus statistics, especially for AI domain
% 2. labeled NER, we focus on low-resource scenario, need to discuss about entity categories
The data statistics of the Reuters News domain~\cite{tjong2003introduction} and the collected five domains are illustrated in Table~\ref{table:datastatistics}. In general, it is easy to collect a large unlabeled corpus for one domain, while for some low-resource domains, the corpus size could be small. As we can see from the statistics of the unlabeled corpus, the size is large for all domains except the AI domain (only a few AI-related pages exist in Wikipedia). Since DAPT experiments usually require a large amount of unlabeled sentences~\cite{wu2020tod,gururangan-etal-2020-dont}, this data scarcity issue introduces a new challenge for the DAPT.
% of fine-tuning pre-trained language models which is under the data scarcity scenario.

We make the size of the training set (from Table~\ref{table:datastatistics}) relatively small since cross-domain NER models are expected to do fast adaptation with a small-scale of target domain data samples. In addition, there are domain-specialized entity types for each domain, resulting in a hierarchical category structure. For example, there are ``politician'' and ``person'' classes, but if a person is a politician, that person should be annotated as a ``politician'' entity, and if not, a ``person'' entity.
Similar cases can be found for ``scientist'' and ``person'', ``organization'' and ``political party'', etc. We believe this hierarchical category structure will bring a challenge to this task since the model needs to better understand the context of inputs and be more robust in recognizing entities.

\begin{figure}[!t]
    \centering
    \includegraphics[scale=0.565]{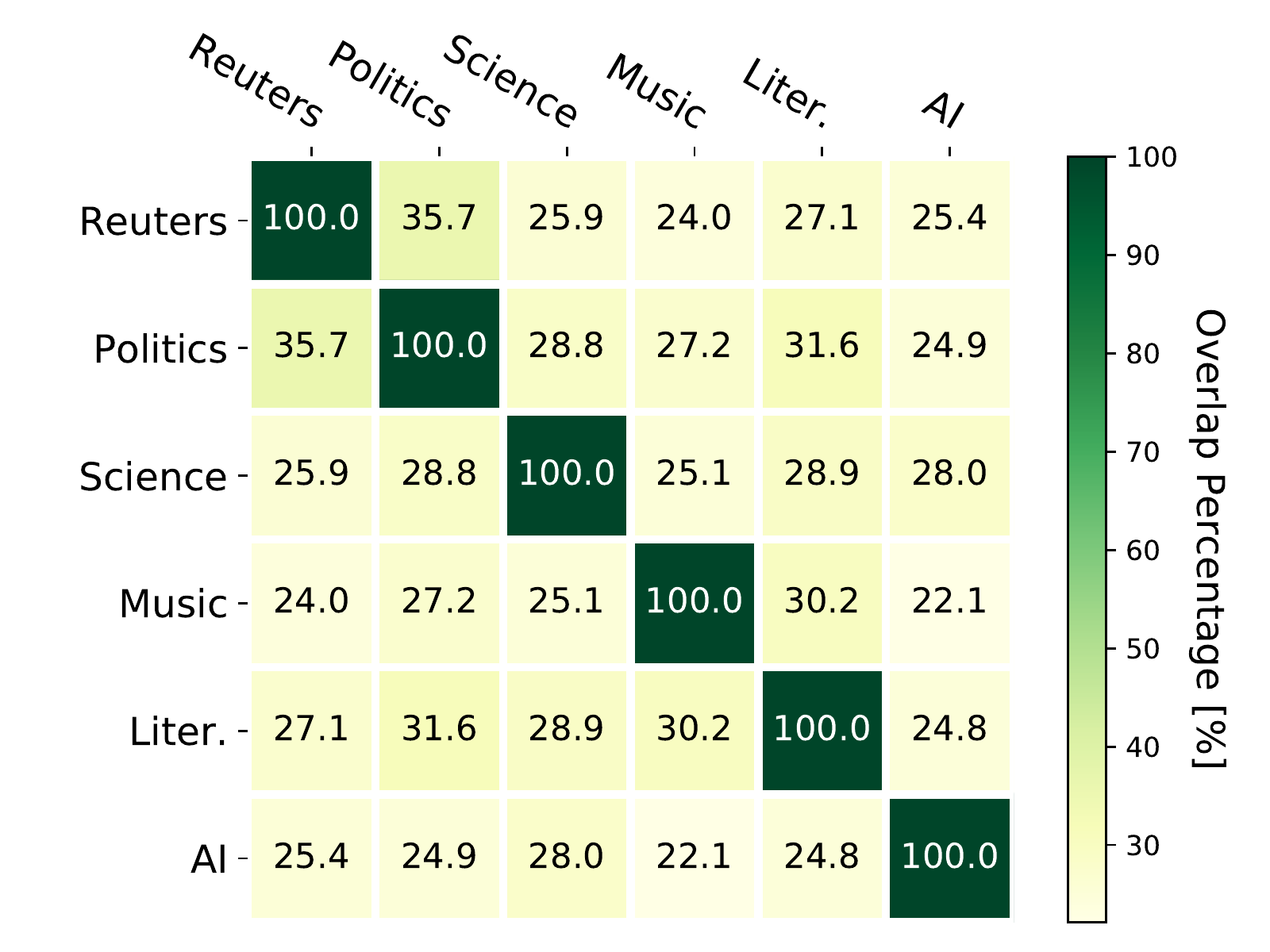}
    \caption{Vocabulary overlaps between domains (\%). Reuters denotes the Reuters News domain, ``Science'' denotes the natural science domain and ``Litera.'' denotes the literature domain.}
    \label{fig:overlap}
\end{figure}

\subsection{Domain Overlap}
The vocabulary overlaps of the NER datasets between domains (including the source domain (Reuters News domain) and the five collected target domains) are shown in Figure~\ref{fig:overlap}. Vocabularies for each domain are created by considering the top 5K most frequent words (excluding stopwords).
We observe that the vocabulary overlaps between domains are generally small, which further illustrates that the overlaps between domains are comparably small and the domains of our collected datasets are diverse. The vocabulary overlaps of the unlabeled corpora between domains are reported in the Appendix.
% Although the overlap between the politics and Reuters domain is slightly larger than others, it is still comparably small. 

\section{Domain-Adaptive Pre-training}
We continue pre-training the language model BERT~\cite{devlin2019bert} on the unlabeled corpus (i.e., DAPT) for the domain adaptation. 
The DAPT is explored in two directions. First, we investigate how different levels of the corpus influences the pre-training.
% how the corpus content influences the domain adaptation ability. 
Second, we explore the effectiveness between token-level and span-level masking in the DAPT.
% language model fine-tuning process for the domain adaptation.

\subsection{Pre-training Corpus}
% domain-level: the largest corpus that we can collect
% entity-level: corpus contain sufficient entities
% task-level: corpus contain domain-specialized entities.
When the size of the domain-related corpus is enormous, continuing to pre-train language models on it would be time-consuming. In addition, there would be noisy and domain-unrelated sentences in the collected corpus which could weaken the effectiveness of the DAPT. Therefore, we investigate whether extracting more indispensable content from the large corpus for pre-training can achieve comparable or even better cross-domain performance.

% First describe how we construct different levels of corpus. We describe usually this kind of corpus can be easily obtain, if not available, we can use NER models to obtain these corpus (later on sections).
% 1. Why it is easy to obtain the corpus
% 2. How we obtain this corpus
We consider three different levels of corpus for pre-training. The first is the \textbf{domain-level} corpus, which is the largest corpus we can collect related to a certain domain. 
The second is the \textbf{entity-level} corpus. It is a subset of the domain-level corpus and made up of sentences having plentiful entities. Practically, it can be extracted from the domain-level corpus based on an entity list. We leverage the entity list in DBpedia Ontology and extract sentences that contain multiple entities to construct the entity-level corpus. 
The third is the \textbf{task-level} corpus, which is explicitly related to the NER task in the target domain. 
% In a real scenario, there would be a set of unlabeled target domain documents required for the text processing and entity recognition, and these documents can be considered as the task-level corpus. 
To construct this corpus, we select sentences having domain-specialized entities existing in the DBpedia Ontology. The size of the task-level corpus is expected to be much smaller than the entity-level corpus. However, its content should be more beneficial than that of the entity-level corpus.

Taking this further, we propose to \textbf{integrate} the entity-level and the task-level corpus. Instead of simply merging these two corpora, we first upsample the task-level corpus (double the size in practice) and then combine it with the entity-level corpus. Hence, models will tend to focus more on the task-level sentences in the DAPT.

% put it in the experimental part
% The data size comparison for aforementioned corpus types is shown in Table xxx.

\subsection{Span-level Pre-training}
Inspired by~\citet{joshi2020spanbert}, we propose to change the token-level masking (MLM) in BERT~\cite{devlin2019bert} into span-level masking for the DAPT. In BERT, MLM first randomly masks 15\% of the tokens in total, and then replaces 80\% of the masked tokens with special tokens (\texttt{[MASK]}), 10\% with random tokens and 10\% with the original tokens. 
We follow the same masking strategy as BERT except the first masking step. In the first step, after the random masking, we move the individual masked index position into its adjacent position that is next to another masked index position in order to produce more masked spans, while we do not touch the continuous masked indices (i.e., masked spans). For example, the randomly masked sentence:

\texttt{Western music's effect would [MASK] to grow within the country [MASK] sphere} \\
\noindent would become 

\texttt{Western music's effect would continue to grow within the [MASK] [MASK] sphere}.

Intuitively, span-level masking provides a more challenging task for pre-trained language models. For example, predicting ``San Francisco'' is much harder than predicting ``San'' given ``Francisco'' as the next word. Hence, the span-level masking can facilitate BERT to better understand the domain text so as to complete the more challenging task.

\begin{table*}[h!]
\renewcommand{\arraystretch}{1.15}
\centering
\resizebox{0.795\textwidth}{!}{
\begin{tabular}{|c|c|c|c|c|c|c|c|c|}
\hline
\textbf{Models}        & \textbf{Masking}                      & \textbf{Corpus}                 & \textbf{Politics} & \textbf{Science} & \textbf{Music} & \textbf{Litera.} & \textbf{AI}    &      \textbf{Average}         \\ \hline \hline
\multicolumn{9}{|l|}{\textbf{\textit{Fine-tune Directly on Target Domains (Directly Fine-tune)}}} \\ \hline
\multirow{8}{*}{BERT-based}     & \multicolumn{2}{c|}{w/o DAPT}               & 66.56    & 63.73   & 66.59 & 59.95      & 50.37   & 61.44              \\ \cline{2-9} 
    & \multirow{4}{*}{Token-level} & Domain-level           & 67.21    & 64.63   & 70.56 & 62.54      & 53.66  & 63.72                     \\ 
&       & Entity-level           & 67.59    & 65.97   & 70.64 & 63.77      & 53.94      & 64.38                 \\ 
&        & Task-level             & 67.30     & 65.04   & 70.37 & 62.10       & 53.19      & 63.60                 \\ 
 & & Integrated             & \textbf{68.83}    & \textbf{66.55}   & \textbf{72.42} & \textbf{63.95}      & \textbf{55.44}      & \textbf{65.44}                 \\ \cline{2-9} 
 & \multirow{3}{*}{Span-level}  & Entity-level           & 68.58    & 66.70    & 71.62 & 64.67      & 55.65  & 65.44                     \\
  &         & Task-level             & 68.37    & 65.84   & 70.66 & 63.85      & 54.48    & 64.64                   \\
 &      & Integrated             & \textbf{70.45}    & \textbf{67.59}   & \textbf{73.39} & \textbf{64.96}      & \textbf{56.36}     & \textbf{66.55}                  \\ \hline \hline
\multicolumn{9}{|l|}{\textbf{\textit{Pre-train on the Source Domain then Fine-tune on Target Domains (Pre-train then Fine-tune)}}} \\ \hline
       \multirow{8}{*}{BERT-based}  & \multicolumn{2}{c|}{w/o DAPT}     & 68.71    & 64.94   & 68.30  & 63.63      & 58.88     & 64.89                  \\ \cline{2-9}
& \multirow{4}{*}{Token-level} & Domain-level           & 69.37    & 66.68   & 72.05 & 65.15      & 61.48   & 66.95                    \\
  &    & Entity-level           & 70.32    & 67.03   & 71.55 & 65.76      & 61.52    & 67.24                   \\       &                                              &            Task-level             & 70.21    & 65.99   & 71.74 & 65.32      & \multicolumn{1}{l|}{60.29} & 66.71  \\   &                                              &      Integrated             & \textbf{71.44}    & \textbf{67.53}   & \textbf{74.02} & \textbf{66.57}      & \multicolumn{1}{l|}{\textbf{61.90}} & \textbf{68.29}  \\ \cline{2-9}                                               & \multirow{3}{*}{Span-level}  & Entity-level           & 71.85    & 68.04   & 73.34 & 66.28      & \multicolumn{1}{l|}{61.66} & 68.23  \\   &     & Task-level             & 70.77    & 67.41   & 73.01 & 66.58      & \multicolumn{1}{l|}{61.68} & 67.89  \\   &       & Integrated             & \textbf{72.05}    & \textbf{68.78}   & \textbf{75.71} & \textbf{69.04}      & \multicolumn{1}{l|}{\textbf{62.56}} & \textbf{69.63}  \\ \hline \hline
\multicolumn{9}{|l|}{\textbf{\textit{Jointly Train on Both Source and Target Domains (Jointly Train)}}} \\ \hline
     \multirow{8}{*}{BERT-based}     & \multicolumn{2}{c|}{w/o DAPT}               & 68.85    & 65.03   & 67.59 & 62.57      & 58.57       & 64.52                \\ \cline{2-9} 
 & \multirow{4}{*}{Token-level} & Domain-level           & 69.49    & 66.37   & 71.94 & 63.74      & 60.53     & 66.41                  \\ 
& \multicolumn{1}{l|}{} &       Entity-level           & 70.01    & 66.55   & 71.51 & 63.35      & 61.29     & 66.54                  \\  
& \multicolumn{1}{l|}{} &     Task-level             & 70.14    & 66.06   & 70.70  & 62.68      & 60.14  & 65.94                     \\  
& \multicolumn{1}{l|}{} &     Integrated             & \textbf{71.09}    & \textbf{67.58}   & \textbf{72.57} & \textbf{64.27}      & \textbf{62.55}       & \textbf{67.61}                \\ \cline{2-9} 
& \multirow{3}{*}{Span-level}  & Entity-level           & 71.90     & 68.04   & 71.98 & 64.23      & 61.63    & 67.55                   \\ 
& \multicolumn{1}{l|}{} &    Task-level             & 71.31    & 67.75   & 71.17 & 63.24      & 60.83      & 66.86                 \\ 
& \multicolumn{1}{l|}{}    &    Integrated             & \textbf{72.76}    & \textbf{68.28}   & \textbf{74.30}  & \textbf{65.18}      & \textbf{63.07}        & \textbf{68.72}               \\ \hline \hline
\multicolumn{9}{|l|}{\textbf{\textit{Baseline Models}}} \\ \hline
\multicolumn{1}{|l|}{BiLSTM-CRF (word)}  &    \multicolumn{1}{c|}{-}       & \multicolumn{1}{c|}{-} & 52.52    & 44.6    & 40.77 & 35.69      & 38.24     & 42.36           \\ 
\multicolumn{1}{|l|}{BiLSTM-CRF (word + char)}   & \multicolumn{1}{c|}{-}       & \multicolumn{1}{c|}{-} & 56.60    & 49.97   & 44.79 & 43.03      & 43.56    & 47.59                   \\ \hline
\multicolumn{1}{|l|}{Coach (word)}      & \multicolumn{1}{c|}{-}       & \multicolumn{1}{c|}{-} & 54.01    & 44.88   & 45.58 & 36.18      & 40.41      & 44.21                 \\
\multicolumn{1}{|l|}{Coach (word + char)}  & \multicolumn{1}{c|}{-}       & \multicolumn{1}{c|}{-} & 61.50     & 52.09   & 51.66 & 48.35      & 45.15    & 51.75                   \\ \hline
\multicolumn{1}{|l|}{\citet{jia2019cross}}          & \multicolumn{1}{c|}{-}       & \multicolumn{1}{c|}{-} & 68.44    & 64.31   & 63.56 & 59.59      & 53.70    & 61.92                   \\ \hline
\multicolumn{1}{|l|}{\citet{jia2020multi}}          & \multicolumn{1}{c|}{-}       & \multicolumn{1}{c|}{-} & 70.56    & 66.42   & 70.52 & 66.96      & 58.28       & 66.55                \\
\multicolumn{1}{|l|}{+ DAPT (Span-level \& Integrated)}          & \multicolumn{1}{c|}{-}       & \multicolumn{1}{c|}{-} & \textbf{71.45}    & \textbf{67.68}   & \textbf{74.19} & \textbf{68.63}      & \textbf{61.64}       & \textbf{68.71}                \\ \hline
\end{tabular}
}
\caption{F1-scores of our proposed methods in three settings and baseline models. Results are averaged over three runs.}
% report variances in the appendix
\label{main_table}
\end{table*}

\section{Experiments}
\subsection{Experimental Settings}
We consider the CoNLL2003 English NER dataset~\cite{tjong2003introduction} from Reuters News, which contains person, location, organization and miscellaneous entity categories, as the source domain and five domains in CrossNER as target domains. Our model is based on BERT~\cite{devlin2019bert} in order to have a fair comparison with the current state-of-the-art model~\cite{jia2020multi}, and we follow~\citet{devlin2019bert} to fine-tune BERT on the NER task. More training details are in the Appendix.

Before training on the source or target domains, we conduct the DAPT on BERT when the unlabeled domain-related corpus is leveraged. Moreover, in the DAPT, different types of unlabeled corpora are investigated (i.e., domain-level, entity-level, task-level and integrated corpora), and different masking strategies are inspected (i.e., token-level and span-level masking).
Then, we carry out three different settings for the domain adaptation, which are described as follows:
\begin{itemize}
    \item We ignore the source domain training samples, and fine-tune BERT directly on the target domain data.
    \item We first pre-train BERT on the source domain data, and then fine-tune it to the target domain samples.
    \item We jointly fine-tune BERT on both source and target domain data samples. Since the size of the data samples in the target domains is smaller than in the source domain, we upsample the target domain data to balance the source and target domain data samples.
\end{itemize}

\subsection{Baseline Models}
We compare our methods to the following baselines:
\begin{itemize}
    \item BiLSTM-CRF~\cite{lample2016neural} incorporate bi-directional LSTM~\cite{hochreiter1997long} and conditional random fields for named entity recognition. We combine source domain data samples and the upsampled target domain data samples to jointly train this model (i.e., the joint training setting mentioned in the experimental settings). We use the word-level embeddings from~\citet{pennington2014glove} and the char-level embeddings from~\citet{hashimoto2017joint}.
    \item \citet{liu-etal-2020-coach} proposed a new framework, Coach, for slot filling and NER domain adaptation. It splits the task into two stages by first detecting the entities and then categorizing the detected entities.
    \item \citet{jia2019cross} integrated language modeling tasks and NER tasks in both source and target domains to perform cross-domain knowledge transfer. We follow their settings and provide the domain-level corpus for the language modeling tasks.
    \item \citet{jia2020multi} proposed a multi-cell compositional LSTM structure based on BERT representations~\cite{devlin2019bert} for domain adaptation, which is the current state-of-the-art cross-domain NER model. 
    % add more baselines (as many as possible)
\end{itemize}

\section{Results \& Analysis}

% analyze domain similarity
\subsection{Corpus Types \& Masking Strategies}
% Corpus Types
From Table~\ref{main_table}, we can see that DAPT using the entity-level or the task-level corpus achieves better or on par results with using the domain-level corpus, while according to the corpus statistics illustrated in Table~\ref{corpus_statistics}, the size of the entity-level corpus is generally around half or less than half that of the domain-level corpus, and the size of the task-level corpus is much smaller than the domain-level corpus. 
We conjecture that the content of the corpus with plentiful entities is more suitable for the NER task's DAPT. In addition, selecting sentences with plentiful entities is able to filter numerous noisy sentences and partial domain-unrelated sentences from the domain corpus. Picking sentences having domain-specialized entities also filters a great many sentences that are not explicitly related to the domain and makes the DAPT more effective and efficient. 
In general, DAPT using the task-level corpus performs slightly worse than using the entity-level corpus. This can be attributed to the large corpus size differences. Furthermore, integrating the entity-level and task-level corpora is able to consistently boost the adaptation performance compared to utilizing other corpus types, although the size of the integrated corpus is still smaller than the domain-level corpus. This is because the integrated corpus ensures the pre-training corpus is relatively large, and in the meantime, focuses on the content that is explicitly related to the NER task in the target domain. 
The results suggest that the corpus content is essential for the DAPT, and \textit{we leave exploring how to extract effective sentences for the DAPT for future work}.
Surprisingly, the DAPT is still effective for the AI domain even though the corpus size in this domain is relatively small, which illustrates that the DAPT is also practically useful in a small corpus setting.

\begin{table*}[!ht]
\renewcommand{\arraystretch}{1.2}
\centering
\resizebox{0.775\textwidth}{!}{
\begin{tabular}{|c|c|c|c|c|c|}
\hline
& \textbf{Politics} & \textbf{Science} & \textbf{Music}   & \textbf{Litera.} & \textbf{AI}     \\ \hline
Domain-level & 177M (1x)  & 99M (1x)  & 195M (1x) & 177M (1x) & 5.2M (1x)  \\ \hline
Entity-level & 67M (0.37x)   & 36M (0.36x)  & 96M (0.49x)  & 87M (0.49x)  & 2.7M (0.52x)  \\ \hline
Task-level   & 16M (0.09x)   & 3.9M (0.04x)  & 26M (0.13x)  & 14M (0.08x)  & 0.2M (0.04x) \\ \hline
Integrated   & 99M (0.56x)  & 44M (0.44x)  & 148M (0.76x) & 115M (0.65x) & 3.1M (0.60x)  \\ \hline
\end{tabular}
}
\caption{Number of tokens of different corpus types. The number in the brackets represents the size ratio between the corresponding corpus and the domain-level corpus.}
\label{corpus_statistics}
\end{table*}

\begin{figure*}[!ht]
\centering
\begin{subfigure}{.33\textwidth}
    \centering
    \includegraphics[scale=0.49]{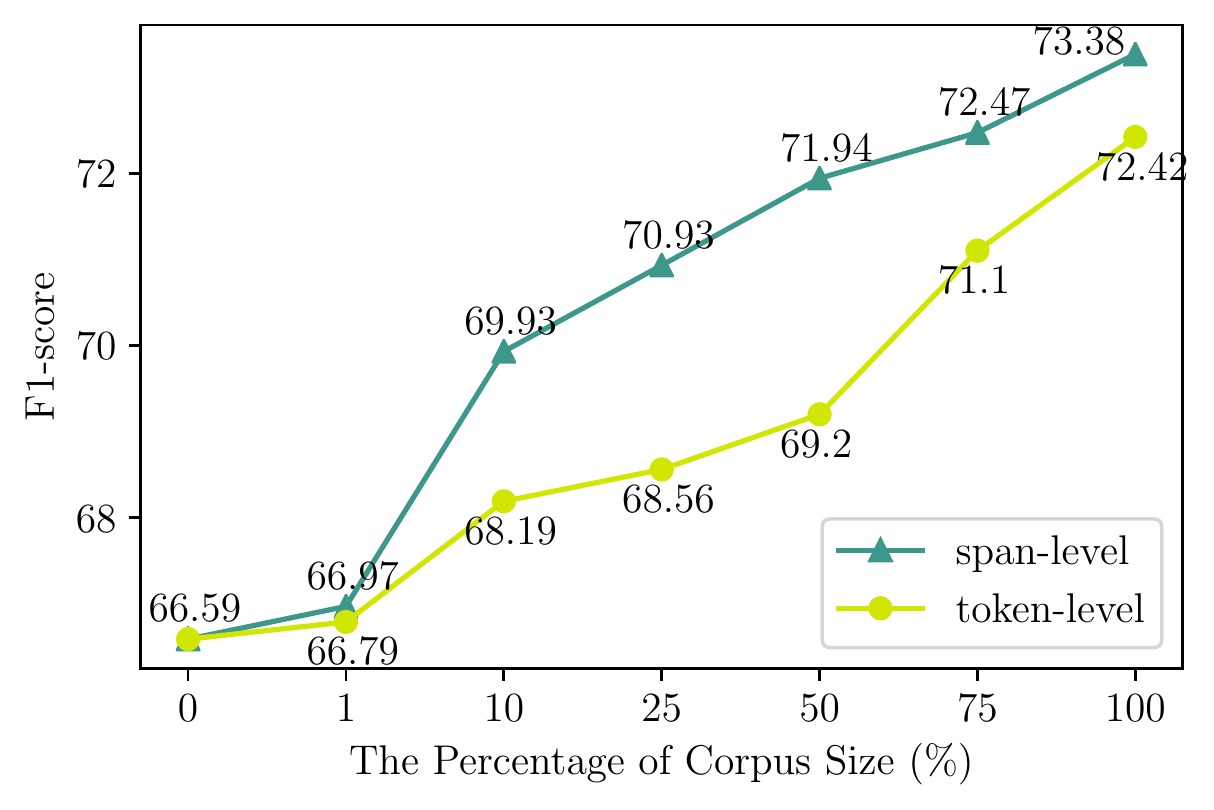}
    \caption{Directly Fine-tune.}
    \label{fig:intent-es}
\end{subfigure}
\begin{subfigure}{.33\textwidth}
    \centering
\includegraphics[scale=0.49]{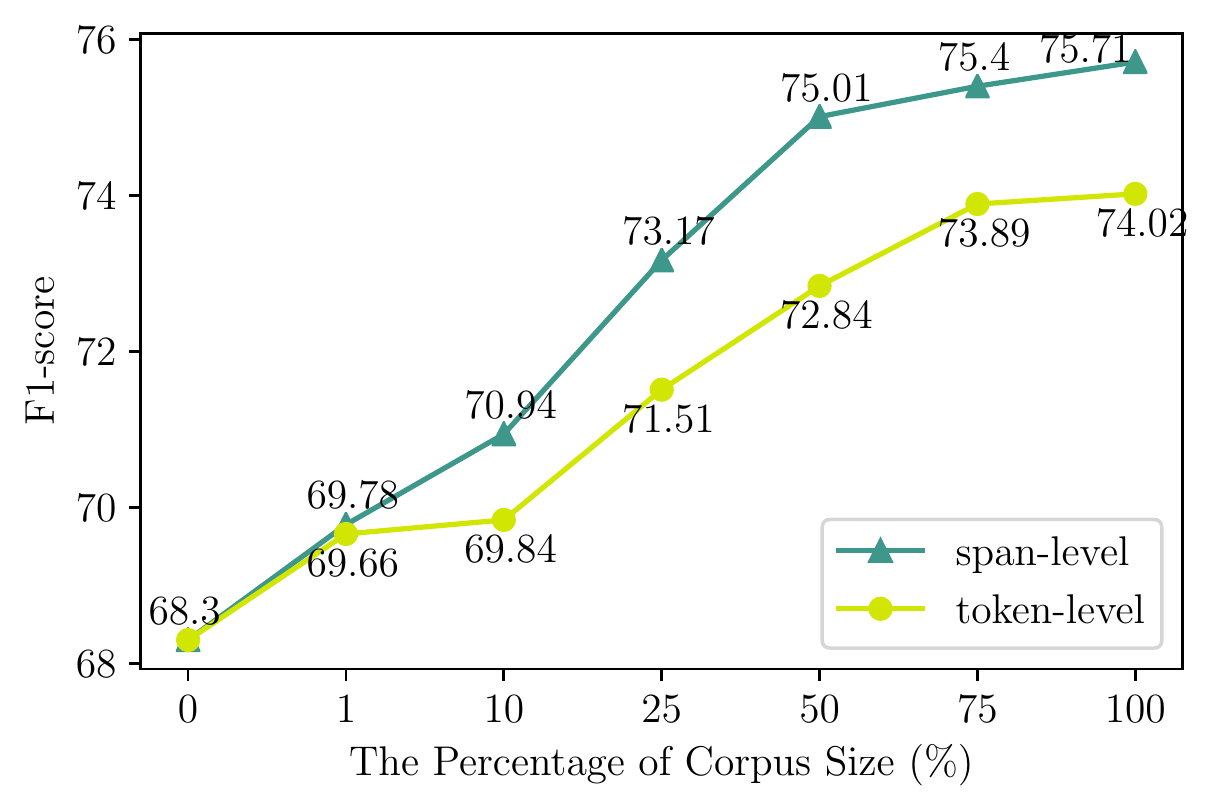}
    \caption{Pre-train then Fine-tune.}
    \label{fig:slot-es}
\end{subfigure}
\begin{subfigure}{.33\textwidth}
    \centering
    \includegraphics[scale=0.49]{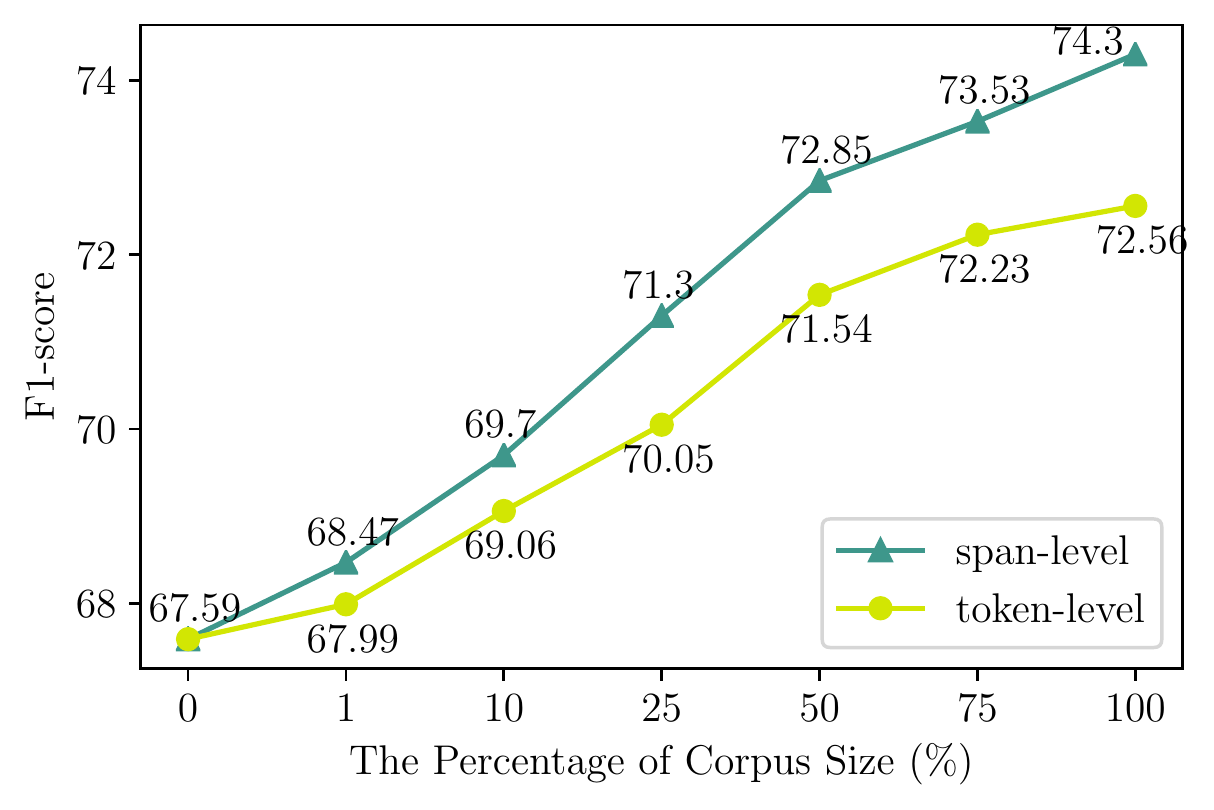}
    \caption{Jointly Train.}
    \label{fig:intent-es-unseen}
\end{subfigure}
\caption{Comparisons among utilizing different percentages of the music domain's integrated corpus and masking strategies in the DAPT for the three settings.}
\label{fig:performance_vs_corpus_size}
\end{figure*}

% Masking Strategies
As we can see from Table~\ref{main_table}, when leveraging the same corpus, the span-level masking consistently outperforms the token-level masking. For example, in the Pre-train then Fine-tune setting, DAPT on the integrated corpus and using span-level masking outperforms that using token-level masking by a 1.34\% F1-score on average. This is because predicting spans is a more challenging task than predicting tokens, forcing the model to better comprehend the domain text and then to possess a more powerful capability to do the downstream tasks.
Moreover, adding DAPT using the span-level masking and the integrated corpus to~\citet{jia2020multi} further improves the F1-score by 2.16\% on average. 
% Although the span-level masking achieves promising results, 
\textit{Nevertheless, we believe that exploring more masking strategies or DAPT methods is worthwhile. We leave this for future work}.

% comparison among three settings
From Table~\ref{main_table}, we can clearly observe the improvements when the source domain data samples are leveraged. For example, compared to the Directly Fine-tune, Pre-train then Fine-tune (w/o DAPT) improves the F1-score by 3.45\% on average, and Jointly Train (w/o DAPT) improves the F1-score by 3.08\% on average.
% pre-training on the source domain NER samples (i.e., the second setting) without DAPT, the F1-score is improved by 3.45\% on average, 
% and when the source domain NER samples are incorporated to jointly train the model (i.e., the third setting) without DAPT, the F1-score is improved by 3.08\% on average. 
We notice that Pre-train then Fine-tune generally leads to better performance than Jointly Train. We speculate that jointly training on both the source and target domains makes it difficult for the model to concentrate on the target domain task, leading to a sub-optimal result, while for the Pre-train then Fine-tune, the model learns the NER task knowledge from the source domain data in the pre-training step and then focuses on the target domain task in the fine-tuning step.
Finally, we can see that our best model can outperform the existing state-of-the-art model in all five domains.
However, the averaged F1-score of the best model is not yet perfect (lower than 70\%), which highlights the need for more advanced cross-domain models.
% in general, is still far from satisfactory compared to the performance of simple LSTM-based models trained on extensive data samples~\cite{ma2016end,chiu2016named,lample2016neural}, which highlights that a more advanced cross-domain NER model is needed.

\subsection{Performance vs. Unlabeled Corpus Size}
Given that a large-scale domain-related corpus might sometimes be unavailable, we investigate the effectiveness of different corpus sizes for the DAPT and explore how the masking strategies will influence the adaptation performance.
% Analysis
% 1. Generally speaking, as the corpus size increases, performance keeps improving. While, we observe that the performance in the setting (b) starts to grow slowly, we conjecture that the performance is difficult to increase when it reaches a certain high level.
% 2. Span-level masking outperforms token-level when the corpus size is large enough. In setting (a), the improvement margin starts to get close when the corpus size is large enough.
As shown in Figure~\ref{fig:performance_vs_corpus_size}, as the size of unlabeled corpus increases, the performance generally keeps improving. 
This implies that the corpus size is generally essential for the DAPT, and within a certain corpus size, the larger the corpus is, the better the domain adaptation performance the DAPT will produce.
Additionally, we notice that in the Pre-train then Fine-tune setting, the improvement becomes comparably less when the percentage reaches 75\% or higher. We conjecture that it is relatively difficult to improve the performance when it reaches a certain amount.

Furthermore, little performance improvements are observed for both the token-level and span-level masking strategies when only a small-scale corpus (1\% of the music integrated corpus, 1.48M) is available. As the corpus size increases, the span-level masking starts to outperform the token-level masking. We notice that in Directly Fine-tune, the performance discrepancy between the token-level and span-level is first increasing and then decreasing. And the performance discrepancies are generally increasing in the other two settings. We hypothesize that the span-level masking can learn the domain text more efficiently since it is a more challenging task, while the token-level masking requires a larger corpus to better understand the domain text.
% It is because that the performance of token-level masking is increasing relatively slow when the corpus is not large enough (the percentage is less then 50\%), while it starts to increase faster when the percentage reaches 75\% or higher. By contrast, the performance of the span-level is increasing fast and then slow as the corpus size increases. 

\begin{figure*}[ht!]
    \centering
    \includegraphics[scale=0.26]{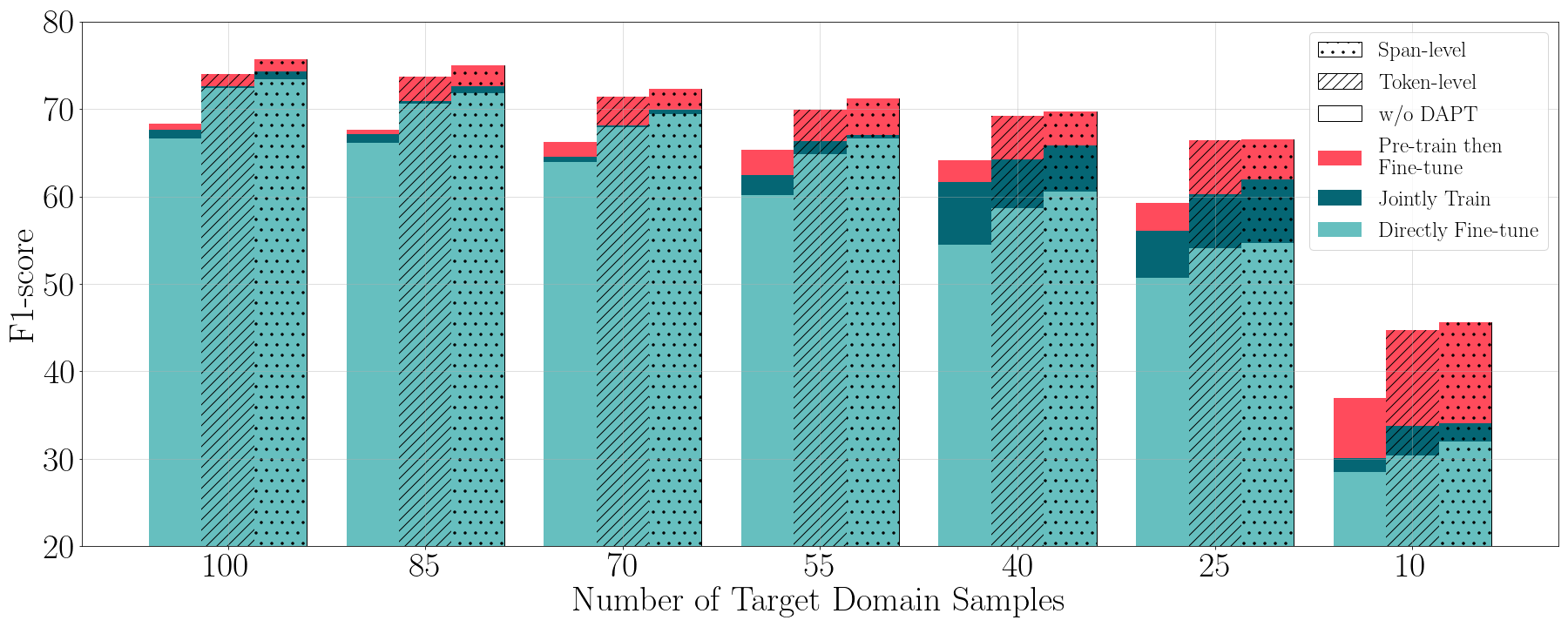}
    \caption{Few-shot F1-scores (averaged over three runs) in the music domain. We use the integrated corpus for the DAPT.}
    \label{fig:fewshot}
\end{figure*}

\begin{table*}[]
\centering
\resizebox{0.935\textwidth}{!}{
\begin{tabular}{|l|c|c|c|c|c|c|c|c|c|c|}
\hline
\multicolumn{1}{|c|}{\textbf{Models}} & \textbf{Genre} & \textbf{Song}  & \textbf{Band}  & \textbf{Album} & \textbf{Artist} & \textbf{Country} & \textbf{Loc.}  & \textbf{Org.}  & \textbf{Per.}  & \textbf{Misc.}  \\ \hline
\multicolumn{11}{|l|}{\textit{\textbf{Directly Fine-tune}}} \\ \hline
w/o DAPT                     & 77.35      & 26.57 & 71.65 & 60.61 & 80.71       & 84.93   & 67.82  & 59.36 & 7.97   & 17.20 \\
Span-level + Integrated      & 78.89      & 42.63 & \textbf{83.03} & 65.93 & \textbf{85.70}       & 83.86   & 77.17 & 69.43 & 10.09 & 19.20 \\ \hline
\multicolumn{11}{|l|}{\textit{\textbf{Pre-train then Fine-tune}}}                                  \\ \hline
w/o DAPT                     & 79.12      & 42.04 & 68.45  & 61.79  & 76.75        & \textbf{88.85}   & 78.47 & 69.70 & 10.15  & \textbf{29.80} \\
Span-level + Integrated      & \textbf{80.82}      & \textbf{58.67} & 82.72 & \textbf{69.28} & 84.58       & 85.61   & \textbf{80.54} & \textbf{75.16} & \textbf{12.59}  & 28.67 \\ \hline
\end{tabular}
}
\caption{F1-scores (averaged over three runs) for the categories in the music domain over the Directly Fine-tune and Pre-train then Fine-tune settings. Span-level+Integrated denotes that the span-level masking and integrated corpus are utilized for the DAPT. ``Loc.'', ``Org.'', ``Per.'' and ``Misc.'' denote ``Location'', ``Organization'', ``Person'' and ``Miscellaneous'', respectively.}
\label{fine-grained}
\end{table*}

\subsection{Performance vs. Target Domain Sample Size}
% 1. In general, the performance drops when the number of target domain samples is reduced, and the task is extremely difficult when the sample number is very few (e.g., 10 samples).
% 2. without source domain, performance drops rapidly when the number of samples is reduced to a certain point. w/ DAPT, performance significantly outperforms w/o DAPT for the other two settings.
% 3. when the number of sample is very little (10 samples), pre-training then fine-tuning significantly outperforms Jointly Train.
From Figure~\ref{fig:fewshot}, we can see that the performance drops when the number of target domain samples is reduced, the span-level pre-training generally outperforms token-level pre-training, and the task becomes extremely difficult when only a few data samples (e.g., 10 samples) in the target domain are available. Interestingly, as the target domain sample size decreases, the advantage of using source domain training samples becomes more significant, for example, Pre-train then Fine-tune outperforms Directly Fine-tune by $\sim$10\% F1 when the sample size is reduced to 40 or lower than 40. This is because these models are able to gain the NER task knowledge from the large amount of source domain examples and then possess the ability to quickly adapt to the target domain.
Additionally, using DAPT significantly improves the performance in the Pre-train then Fine-tune setting when target domain samples are scarce (e.g., 10 samples). This can be attributed to the boost in the domain adaptation ability made by the DAPT, which allows the model to quickly learn the NER task in the target domain. Furthermore, we notice that with the decreasing of the sample size, the performance discrepancy between Pre-train then Fine-tune and Jointly Train is getting larger. We speculate that in the Jointly Train setting, the models focus on the NER task in both the source and target domains. This makes the models tend to ignore the target domain when the sample size is too small, while for the Pre-train then Fine-tune setting, the models can focus on the target domain in the fine-tuning stage to ensure the good performance in the target domain.

\subsection{Fine-grained Comparison}
In this section, we further explore the effectiveness of the DAPT and leveraging NER samples in the source domain. As shown in Table~\ref{fine-grained}, the performance is improved on almost all categories when the DAPT or the source domain NER samples are utilized. We observe that using source domain NER data might hurt the performance in some domain-specialized entity categories, such as ``artist'' (``musical artist'') and ``band''. This is because that since ``artist'' is a subcategory of ``person'' and models pre-trained on the source domain tend to classify artists as ``person'' entities. Similarly, ``band'' is a subcategory of ``organization'', which leads to the same misclassification issue after the source domain pre-training.
When the DAPT is used, the performance on some domain-specialized entity categories is greatly improved (e.g., ``song'', ``band'' and ``album'').
We notice that the performance on the ``person'' entity is relatively low compared to other categories. It is because that the hierarchical category structure could cause models to be confused between ``artist'' and ``person'' entities, and we find out that 84.81\% of ``person'' entities are misclassified as ``artist'' in the best model we have.

% Interesting, the DAPT boosts the performance on the Person, and Miscellaneous categories in the Directly Fine-tune setting, while makes the performance drop in the Pre-train then Fine-tune setting, especially for the Person category. We conjecture that the discrepancy between source and target domains could cause the DAPT on the music domain to hurt the pre-training on the source domain in some non-domain-specialized categories.

\section{Conclusion}
In this paper, we introduce CrossNER, a humanly-annotated NER dataset spanning over five diverse domains with specialized entity categories for each domain. In addition, we collect the corresponding domain-related corpora for the study of DAPT. A set of benchmark results of existing strong NER models is reported. Moreover, we conduct comprehensive experiments and analyses in terms of the size of the domain-related corpus and different pre-training strategies in the DAPT for the cross-domain NER task, and our proposed method consistently outperforms existing baselines. Nevertheless, the performance of our best model is not yet perfect, especially when the number of target domain training samples is limited. We hope that our dataset will facilitate further research in the NER domain adaptation field.

\section*{Acknowledgments}
We want to thank Genta Indra Winata for providing insightful comments for this research project. We also want to thank the anonymous reviewers for their constructive feedback. This work is partially funded by ITF/319/16FP and MRP/055/18 of the Innovation Technology Commission, the Hong Kong SAR Government.

\bibliography{aaai2021}

\appendix
\section{The CrossNER Dataset}
\subsection{Annotator Training}
We gather the NER experts who are familiar with the NER task and the annotation rules. We train the annotators before he/she starts annotating. For each domain, we first give the annotator the annotation instruction and 100 annotated examples (annotated by the NER experts) and ask them to check the possible annotation errors.
After the checking process, the NER experts will inspect the results and tell the annotators the mistakes they made in the checking stage. Hence, in this process, the annotators are able to learn how to annotate the NER samples for specific domains.

\subsection{Annotation Instructions}
We split the annotation instructions into two parts, namely, general instructions and domain-specific instructions. We describe the instructions as follows:

\subsubsection{General Instructions}
Each data sample requires two well-trained NER annotators to annotate it and one NER expert to double check it and give final labels. The annotation process consists of three steps. First, one annotator needs to detect and categorize the entities in the given sentences. Second, the other annotator checks the annotations made by the first annotator, makes markings if he/she thinks that the annotations \textbf{could} be wrong and gives another annotation. Finally, the expert first goes through the annotations again and checks for possible mistakes, and then makes the final decision for disagreements between the first two annotators. Notice that we mark the tokens with hyperlinks in Wikipedia, and it is highly likely that these tokens are named entities. When one entity contains another entity, we should give labels to the entity with a larger span. For example, ``Fellow of the Royal Society'' is an entity (a award entity), while ``Royal Society'' is another entity (an organization entity), we should annotate ``Fellow of the Royal Society'' instead of ``Royal Society''.
The requirements for the annotators in different annotation stages are as follows:
\begin{itemize}
    \item If you are in the first annotation stage, you need to detect and categorize the entities carefully, and check the pre-labeled entities (annotated by DBpedia Ontology described in the main paper) are correct or not, and give the correct annotations if you think the labels are wrong. Notice that the pre-label entity might not be an entity, and some tokens not labeled as entities could be entities.
    \item If you are in the second annotation stage, you need to focus on looking for the possible annotations mistakes made by the first annotator, and give another annotation if you think the labels \textbf{could} be wrong. Additionally, you need to detect and categorize the entities that are missed by the first annotator.
    \item If you are in the third annotation stage (only applicable for NER experts), you need to carefully go through the annotations and correct the possible mistakes, and in the meantime, you need to check the corrected annotations made by the second annotator and then makes final decisions for the disagreements between the two annotators. If you are unsure about the annotations, you need to confer with the two annotators.
\end{itemize}

\subsubsection{Domain-Specific Instructions}
We list the annotation details for the five domains, namely, politics, natural science, music, literature, and artificial intelligence.
\begin{itemize}
    \item \textbf{Politics:} The entity category list for this domain is \{person, organization, location, politician, political party, election, event, country, miscellaneous\}. The annotation rules for the abovementioned entity categories are as follows:
    \begin{itemize}
        \item \textit{Person:} The name of a person should be annotated as a person entity.
        \item \textit{Politician:} The politician entity. If a person entity is a politician, you should label this person as a politician entity instead of a person entity.
        \item \textit{Location:} The location entity, including place, bridge, city, county and etc.
        \item \textit{Country:} The country entity.
        \item \textit{Event:} The event entity, which includes festival, war, summit, Campaign, and etc.
        \item \textit{Election:} The election entity. If an event entity is an election event, you should label it as an election entity instead of an event entity.
        \item \textit{Organization:} The organization entity.
        \item \textit{political party:} The political party entity. If an organization entity is a political party, you should label it as a political entity instead of an organization entity.
        \item \textit{Miscellaneous:} An entity needs to be classified as the miscellaneous type if it does not belong to any other category.
    \end{itemize}
    Note that the annotation rules for some general entity categories (i.e., person, location, organization, event, country, miscellaneous) are the same as those in the other domains, and we do not put the annotation rules for these entity categories for the other domains.
    \item \textbf{Natural Science:} This domain contains the area of biology, chemistry, and astrophysics. The entity category list for this domain is \{scientist, person, university, organization, country, location, discipline, enzyme, protein, chemical element, chemical compound, astronomical object, academic journal, event, theory, award, miscellaneous\}.\footnote{Since data samples come from Wikipedia instead of academic papers from the natural science field, the entities are generally not difficult to categorize by annotators that are not working on these areas.} The annotation rules for the abovementioned entity categories are as follows:
    \begin{itemize}
        \item \textit{University:} The university entity.
        \item \textit{Discipline:} The discipline entity. It contains the areas and subareas of biology, chemistry and astrophysics, such as quantum chemistry.
        \item \textit{Theory:} The theory entity. It includes law and theory entities, such as ptolemaic planetary theories.
        \item \textit{Award:} The award entity.
        \item \textit{Scientist:} If a person entity is a scientist, you should label this person as a scientist entity instead of a person entity.
        \item \textit{Protein:} The protein entity.
        \item \textit{Enzyme:} Notice that enzyme is a special type of protein. Hence, if a protein entity is an enzyme, you should label this protein as an enzyme entity instead of a protein entity.
        \item \textit{Chemical element:} The chemical element entity. Basically, this category contains the chemical elements from the periodic table.
        \item \textit{Chemical compound:} The chemical compound entity. If a chemical compound entity do not belong to protein or enzyme, you should label it as a chemical compound entity.
        \item \textit{Astronomical object:} The astronomical object entity. 
        \item \textit{Academic journal:} The academic journal entity.
    \end{itemize}
    \item \textbf{Music:} The entity category list for this domain is \{music genre, song, band, album, musical artist, musical instrument, award, event, country, location, organization, person, miscellaneous\}. The annotation rules for the abovementioned entity categories are as follows:
    \begin{itemize}
        \item \textit{Music genre:} The music genre entity, such as country music, folk music and jazz.
        \item \textit{Song:} The song entity.
        \item \textit{Band:} The band entity. If an organization belongs to a band, you should label it as a band entity instead of an organization entity.
        \item \textit{Album:} The album entity.
        \item \textit{Musical artist:} The musical artist entity. It a person is working on the music area (e.g., he/she is a singer, composer, or songwriter), you should label it as a musical artist entity instead of a person entity.
        \item \textit{Musical instrument:} The musical instrument entity, such as piano.
    \end{itemize}
    \item \textbf{Literature:} The entity category list for this domain is \{book, writer, award, poem, event, magazine, literary genre, person, location, organization, country, miscellaneous\}. The annotation rules for the abovementioned entity categories are as follows:
    \begin{itemize}
        \item \textit{Book:} The book entity.
        \item \textit{Poem:} The poem entity.
        \item \textit{Writer:} The writer entity. If a person is working on literature (including writer, novelist, scriptwriter, poet, and etc), you should label it as a writer entity instead of a person entity.
        \item \textit{Magazine:} The magazine that publishes articles as well as any other literature work.
        \item \textit{Literary genre:} The literary genre entity, such as novel and science fiction.
    \end{itemize}
    \item \textbf{Artificial Intelligence:} The entity category list for this domain is \{field, task, product, algorithm, researcher, metrics, university, country, person, organization, location, programming language, conference, miscellaneous\}. The annotators for this domain are all working on the AI area. The annotation rules for the abovementioned entity categories are as follows:
    \begin{itemize}
        \item \textit{Researcher:} The researcher entity. If a person is working on research (including professor, Ph.D. student, researcher in companies, and etc), you should label it as a researcher entity instead of a person entity.
        \item \textit{Field:} The research field entity, such as machine learning, deep learning, and natural language processing.
        \item \textit{Task:} The specific task entity in the research field, such as machine translation and object detection.
        \item \textit{Product:} The product entity that includes the product (e.g., a certain kind of robot like Pepper), system (e.g., facial recognition system) and toolkit (e.g., Tensorflow and PyTorch)
        \item \textit{Algorithm:} The algorithm entity. It contains algorithms (e.g., decision trees) and models (e.g, CNN and LSTM).
        \item \textit{Metrics:} The evaluation metrics, such as F1-score.
        \item \textit{Programming Language:} The programming language, such as Java and Python.
        \item \textit{Conference:} The conference entity. It contains conference and journal entities.
    \end{itemize}
\end{itemize}

\begin{table}[]
\renewcommand{\arraystretch}{1.3}
\centering
\resizebox{0.48\textwidth}{!}{
\begin{tabular}{|c|c|c|c|}
\hline
\textbf{Politics}        & \textbf{Train}         & \textbf{Dev}            & \textbf{Test}          \\ \hline
person          & 14 (1.07\%)   & 286 (8.21\%)   & 354 (8.41\%)  \\ \hline
location        & 297 (22.78\%) & 211 (6.06\%)   & 597 (14.19\%) \\ \hline
organization    & 152 (11.66\%) & 431 (12.37\%)  & 513 (12.19\%) \\ \hline
country         & 62 (4.75\%)   & 183 (5.26\%)   & 418 (9.93\%)  \\ \hline
event           & 22 (1.69\%)   & 186 (5.34\%)   & 195 (4.64\%)  \\ \hline
political party & 195 (14.95\%) & 1053 (30.24\%) & 953 (22.65\%) \\ \hline
politician      & 359 (27.53\%) & 411 (11.80\%)  & 485 (11.53\%) \\ \hline
election        & 123 (9.43\%)  & 528 (15.16\%)  & 434 (10.32\%) \\ \hline
miscellaneous   & 80 (6.13\%)   & 193 (5.54\%)   & 258 (6.13\%)  \\ \hline
\end{tabular}
}
\caption{The number of named entities and percentages for each category in the political domain.}
\label{politics_statistics}
\end{table}

\begin{table}[]
\renewcommand{\arraystretch}{1.3}
\centering
\resizebox{0.48\textwidth}{!}{
\begin{tabular}{|c|c|c|c|}
\hline
\textbf{Natural Science}     & \textbf{Train}         & \textbf{Dev}           & \textbf{Test}          \\ \hline
person              & 97 (9.03\%)   & 153 (6.03\%)  & 230 (7.46\%)  \\ \hline
location            & 65 (6.05\%)   & 142 (5.60\%)  & 169 (5.48\%)  \\ \hline
organization        & 118 (10.99\%) & 278 (10.96\%) & 272 (8.82\%)  \\ \hline
country             & 9 (0.84\%)    & 28 (1.10\%)   & 27 (0.88\%)   \\ \hline
event               & 20 (1.86\%)   & 35 (1.38\%)   & 30 (0.97\%)   \\ \hline
award               & 56 (5.21\%)   & 140 (5.52\%)  & 145 (4.70\%)  \\ \hline
scientist           & 127 (11.82\%) & 328 (12.93\%) & 471 (15.27\%) \\ \hline
chemical compound   & 72 (6.70\%)   & 193 (7.61\%)  & 160 (5.18\%)  \\ \hline
chemical element    & 16 (1.49\%)   & 23 (0.91\%)   & 85 (2.76\%)   \\ \hline
theory              & 8 (0.74\%)    & 7 (0.28\%)    & 20 (0.65\%)   \\ \hline
protein             & 69 (6.42\%)   & 101 (3.98\%)  & 156 (5.06\%)  \\ \hline
enzyme              & 72 (2.05\%)   & 48 (1.89\%)   & 80 (2.59\%)   \\ \hline
university          & 56 (5.21\%)   & 138 (5.44\%)  & 130 (4.21\%)  \\ \hline
astronomical object & 121 (11.27\%) & 372 (14.67\%) & 337 (10.92\%) \\ \hline
discipline          & 22 (2.05\%)   & 41 (1.61\%)   & 73 (2.37\%)   \\ \hline
academic journal    & 28 (2.61\%)   & 149 (5.88\%)  & 180 (5.83\%)  \\ \hline
miscellaneous       & 168 (15.64\%) & 360 (14.20\%) & 520 (16.86\%) \\ \hline
\end{tabular}
}
\caption{The number of named entities and percentages for each category in the natural science domain.}
\label{science_statistics}
\end{table}

\begin{table}[]
\renewcommand{\arraystretch}{1.3}
\centering
\resizebox{0.48\textwidth}{!}{
\begin{tabular}{|c|c|c|c|}
\hline
\textbf{Music}              & \textbf{Train}         & \textbf{Dev}           & \textbf{Test}          \\ \hline
person             & 15 (2.31\%)   & 65 (2.42\%)   & 79 (2.37\%)   \\ \hline
location           & 30 (4.63\%)   & 187 (6.98\%)  & 343 (10.29\%) \\ \hline
organization       & 36 (5.56\%)   & 137 (5.12\%)  & 207 (6.21\%)  \\ \hline
country            & 27 (4.17\%)   & 130 (4.85\%)  & 160 (4.80\%)  \\ \hline
award              & 69 (10.65\%)  & 228 (8.51\%)  & 260 (7.80\%)  \\ \hline
event              & 12 (1.86\%)   & 72 (2.69\%)   & 69 (2.07\%)   \\ \hline
band               & 125 (19.29\%) & 402 (15.01\%) & 462 (13.86\%) \\ \hline
album              & 91 (14.04\%)  & 282 (10.53\%) & 332 (9.96\%)  \\ \hline
musical instrument & 2 (0.31\%)    & 42 (1.57\%)   & 22 (0.66\%)   \\ \hline
musical artist     & 104 (16.05\%) & 472 (17.63\%) & 515 (15.45\%) \\ \hline
music genre        & 88 (13.58\%)  & 360 (13.44\%) & 488 (14.64\%) \\ \hline
song               & 27 (4.17\%)   & 177 (6.61\%)  & 227 (6.81\%)  \\ \hline
miscellaneous      & 22 (3.40\%)   & 124 (4.63\%)  & 170 (5.10\%)  \\ \hline
\end{tabular}
}
\caption{The number of named entities and percentages for each category in the music domain.}
\label{music_statistics}
\end{table}

\begin{table}[ht!]
\renewcommand{\arraystretch}{1.3}
\centering
\resizebox{0.48\textwidth}{!}{
\begin{tabular}{|c|c|c|c|}
\hline
\textbf{Literature}    & \textbf{Train}         & \textbf{Dev}           & \textbf{Test}          \\ \hline
person        & 48 (8.87\%)   & 194 (9.12\%)  & 175 (7.72\%)  \\ \hline
location      & 25 (4.62\%)   & 110 (5.17\%)  & 99 (4.37\%)   \\ \hline
organization  & 18 (3.33\%)   & 115 (5.40\%)  & 110 (4.85\%)  \\ \hline
country       & 25 (4.62\%)   & 127 (5.97\%)  & 101 (4.45\%)  \\ \hline
award         & 34 (6.28\%)   & 124 (5.83\%)  & 141 (6.22\%)  \\ \hline
event         & 10 (1.85\%)   & 65 (3.05\%)   & 45 (1.98\%)   \\ \hline
writer        & 133 (24.58\%) & 544 (25.56\%) & 567 (25.02\%) \\ \hline
book          & 91 (16.82\%)  & 336 (15.78\%) & 418 (18.45\%) \\ \hline
literary genre & 36 (6.65\%)   & 152 (7.14\%)  & 194 (8.56\%)  \\ \hline
poem          & 47 (8.68\%)   & 79 (3.17\%)   & 120 (5.30\%)  \\ \hline
magazine      & 16 (2.96\%)   & 71 (3.34\%)   & 57 (2.52\%)   \\ \hline
miscellaneous & 58 (10.72\%)  & 211 (9.92\%)  & 239 (10.55\%) \\ \hline
\end{tabular}
}
\caption{The number of named entities and percentages for each category in the literature domain.}
\label{literature_statistics}
\end{table}

\begin{table}[ht!]
\renewcommand{\arraystretch}{1.3}
\centering
\resizebox{0.48\textwidth}{!}{
\begin{tabular}{|c|c|c|c|}
\hline
\textbf{Artificial Intelligence} & \textbf{Train}        & \textbf{Dev}           & \textbf{Test}          \\ \hline
person                  & 6 (1.13\%)   & 41 (2.65\%)   & 67 (3.71\%)   \\ \hline
location                & 7 (1.32\%)   & 42 (2.71\%)   & 38 (2.11\%)   \\ \hline
organization            & 48 (9.02\%)  & 94 (6.08\%)   & 145 (8.03\%)  \\ \hline
country                 & 29 (5.45\%)  & 44 (2.84\%)   & 44 (2.44\%)   \\ \hline
researcher              & 54 (10.15\%) & 142 (9.18\%)  & 160 (8.86\%)  \\ \hline
product                 & 58 (10.90\%) & 173 (11.18\%) & 198 (10.97\%) \\ \hline
field                   & 39 (7.33\%)  & 177 (11.44\%) & 205 (11.36\%) \\ \hline
task                    & 60 (11.28\%) & 146 (9.44\%)  & 218 (12.08\%) \\ \hline
university              & 32 (6.02\%)  & 59 (3.81\%)   & 28 (1.55\%)   \\ \hline
programming language    & 25 (4.70\%)  & 35 (2.26\%)   & 60 (3.32\%)   \\ \hline
algorithm               & 80 (15.04\%) & 178 (11.51\%) & 177 (9.81\%)  \\ \hline
metrics                 & 27 (5.08\%)  & 147 (9.50\%)  & 191 (10.58\%) \\ \hline
conference              & 24 (4.51\%)  & 89 (5.75\%)   & 93 (5.15\%)   \\ \hline
miscellaneous           & 43 (8.08\%)  & 180 (11.64\%) & 181 (10.03\%) \\ \hline
\end{tabular}
}
\caption{The number of named entities and percentages for each category in the artificial intelligence domain.}
\label{ai_statistics}
\end{table}

\subsection{Data Statistics for Each Entity Category}
The data statistics for each category for the five domains is illustrated in Table~\ref{politics_statistics}, Table~\ref{science_statistics}, Table~\ref{music_statistics}, Table~\ref{literature_statistics}, and Table~\ref{ai_statistics}.

\begin{table*}[]
\renewcommand{\arraystretch}{1.5}
\centering
\resizebox{0.99\textwidth}{!}{
\begin{tabular}{|c|c|l|l|l|l|l|}
\hline
\textbf{Domains}        & \multicolumn{6}{c|}{\textbf{Examples}}                                                                                                                                                                                                                                                                     \\ \hline
Politics                & \multicolumn{6}{c|}{\begin{tabular}[c]{@{}c@{}}In the subsequent election, \underline{Hugo Chávez} (politician) 's political party, the \underline{United Socialist Party of Venezuela} (political party) \\ drew 48\% of the votes overall.\end{tabular}}                                                                         \\ \hline
Natural Science         & \multicolumn{6}{c|}{\begin{tabular}[c]{@{}c@{}} \underline{Mars} (astronomical object) has four known co-orbital asteroids, such as \underline{5261 Eureka} (astronomical object), \\ all at the \underline{Lagrangian points} (miscellaneous).\end{tabular}}                                                                                   \\ \hline
Music                   & \multicolumn{6}{c|}{\begin{tabular}[c]{@{}c@{}} \underline{House of Pain} (band) abruptly broke up in 1996 after the release of their third album, \underline{Truth Crushed to Earth Shall Rise Again} (album), \\ which featured guest appearance by rappers \underline{Sadat X} (musical artist) of \underline{Brand Nubian} (band), etc.\end{tabular}}   \\ \hline
Literature              & \multicolumn{6}{c|}{\begin{tabular}[c]{@{}c@{}} \underline{Charles} (writer) spent outdoors, but also read voraciously, including the \underline{picaresque novels} (literary genre) of \underline{Tobias Smollett} (writer) and \\ \underline{Henry Fielding} (writer), as well as \underline{Robinson Crusoe} (book) and \underline{Gil Blas} (book).\end{tabular}}               \\ \hline
Artificial Intelligence & \multicolumn{6}{c|}{\begin{tabular}[c]{@{}c@{}} The \underline{gradient decent} (algorithm) can take many iterations to compute a \underline{local minimum} (miscellaneous) with a required \underline{accuracy} (metrics), \\ if the \underline{curvature} (miscellaneous) in different directions is very different for the given function.\end{tabular}} \\ \hline
\end{tabular}
}
\caption{Examples for the five domains. We underline the annotated entities with the corresponding annotations in the brackets.}
\label{example}
\end{table*}

\subsection{Examples of Labeled Data}
NER data examples for the five domains are shown in Table~\ref{example}.

\begin{figure}[!t]
    \centering
    \includegraphics[scale=0.565]{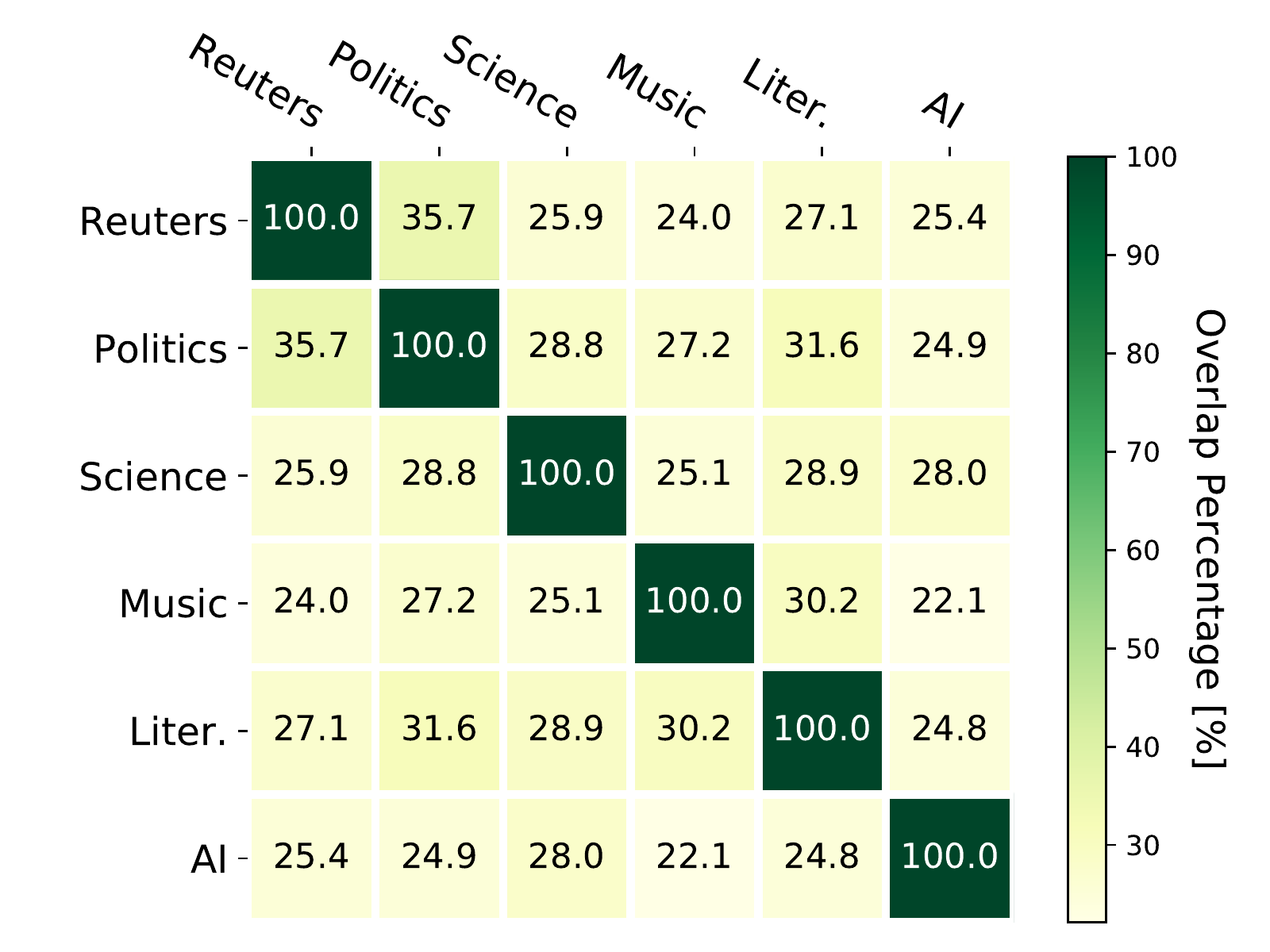}
    \caption{Vocabulary overlaps of the unlabeled domain-related corpora between domains (\%). Reuters denotes the Reuters News domain, ``Science'' denotes the natural science domain, ``Litera.'' denotes the literature domain, and ``AI'' denotes the artificial intelligence domain.}
    \label{fig:overlap}
\end{figure}

\subsection{Vocabulary Overlaps of Unlabeled Corpora}
The vocabulary overlaps of the unlabeled domain-related corpora between domains are shown in Figure~\ref{fig:overlap}. The Reuters News corpus is taken from~\citet{jia2019cross}. For each domain, we sample 150K paragraphs from the domain-related corpus and create the vocabulary by considering the top 50K most frequent words (excluding stopwords). We can see that the overlaps in the unlabeled corpus are generally larger than those in the NER datasets. Since the large corpus size is large, it will contain more frequent words that overlap with those in other domains. In addition, except the vocabulary overlaps between politics and literature domains, and music and literature domains (above 60\%), the overlaps for other domain pairs are still comparably small.

\section{Training Details}
We perform the domain-adaptive pre-training (DAPT) for 15 epochs on the pre-trained corpus. We add a linear layer on top of BERT~\cite{devlin2019bert} and then fine-tune the whole model on the NER task. We select the best hyper-parameters by searching a combination of batch size and learning rate with the following range: batch size \{16, 32\} and learning rate \{$1\times10^{-5}$, $3\times10^{-5}$, $5\times10^{-5}$\}.
In the Directly Fine-tune and Pre-train then Fine-tune settings, we use batch size 16 and learning rate $5\times10^{-5}$, while in the Jointly Train setting, we use batch size 16 and learning rate $1\times10^{-5}$.
Additionally, we upsample the target domain data in the Jointly Train setting to balance the data samples between the source and target domains. Given that the number of training samples in the source domain is around 100 times larger than the target domain samples, the number of times we multiply the target domain data is searched with the range \{10, 50, 100, 150, 200\}, and we find that 100 is generally suitable for all domains. We use F1-score to evaluate the models. It is commonly used evaluation metrics for the NER models~\cite{ma2016end,lample2016neural,chiu2016named}. We use the early stop strategy and select the model based on the performance on the development set of the target domain. 
All the experiments are running on GTX 1080 Ti. We will release our collected labeled NER datasets, unlabeled domain-related corpora, different levels of unlabeled corpora, and all model checkpoints to catalyze the research in the cross-domain NER area.

% \usepackage{multirow}
\begin{table*}[ht!]
\renewcommand{\arraystretch}{1.2}
\centering
\resizebox{0.89\textwidth}{!}{
\begin{tabular}{|c|c|c|c|c|c|c|c|c|}
\hline
\textbf{Models}                      & \textbf{Setting}                                                                            & \textbf{Corpus}       & \textbf{Politics} & \textbf{Science} & \textbf{Music} & \textbf{Litera.} & \textbf{AI}    & \textbf{Average} \\ \hline
\multicolumn{9}{|l|}{\textit{\textbf{Fine-tune Directly on Target Domains (Directly Fine-tune)}}}                                                                                                          \\ \hline
\multirow{7}{*}{BERT-based} & \multicolumn{2}{c|}{w/o DAPT}                                                                     & 66.56    & 63.73   & 66.59 & 59.95   & 50.37 & 61.44   \\ \cline{2-9} 
                            & \multirow{3}{*}{\begin{tabular}[c]{@{}c@{}}DBpedia \\ Ontology-based\end{tabular}} & entity-level & 68.58    & 66.7    & 71.62 & 64.67   & 55.65 & 65.44  \\
                            &                                                                                    & task-level   & 68.37    & 65.84   & 70.66 & 63.85   & 54.48 & 64.64   \\
                            &                                                                                    & integrated   & \textbf{70.45}    & \textbf{67.59}   & \textbf{73.39} & 64.96   & 56.36 & \textbf{66.55}   \\ \cline{2-9} 
                            & \multirow{3}{*}{\begin{tabular}[c]{@{}c@{}}NER \\ Model-based\end{tabular}}        & entity-level & 68.64    & 66.89   & 71.76 & 63.33   & 55.59 & 65.24   \\
                            &                                                                                    & task-level   & 68.48    & 65.34   & 70.80  & 63.05   & 52.77 & 64.09   \\
                            &                                                                                    & integrated   & 69.13    & 67.16   & 73.16 & \textbf{66.15}   & \textbf{56.61} & 66.44   \\ \hline
\multicolumn{9}{|l|}{\textit{\textbf{Pre-train on the Source Domain then Fine-tune on Target Domains (Pre-train then Fine-tune)}}}                                                                         \\ \hline
\multirow{7}{*}{BERT-based} & \multicolumn{2}{c|}{w/o DAPT}                                                                     & 68.71    & 64.94   & 68.3  & 63.63   & 58.88 & 64.89  \\ \cline{2-9} 
                            & \multirow{3}{*}{\begin{tabular}[c]{@{}c@{}}DBpedia \\ Ontology-based\end{tabular}} & entity-level & 71.85    & 68.04   & 73.34 & 66.28   & 61.66 & 68.23  \\
                            &                                                                                    & task-level   & 70.77    & 67.41   & 73.01 & 66.58   & 61.68 & 67.89   \\
                            &                                                                                    & integrated   & 72.05    & 68.78   & 75.71 & \textbf{69.04}   & \textbf{62.56} & 69.63  \\ \cline{2-9} 
                            & \multirow{3}{*}{\begin{tabular}[c]{@{}c@{}}NER \\ Model-based\end{tabular}}        & entity-level & 72.16    & 68.53   & 73.78 & 66.69   & 61.25 & 68.48   \\
                            &                                                                                    & task-level   & 71.98    & 67.43   & 72.92 & 65.96   & 60.02 & 67.66   \\
                            &                                                                                    & integrated   & \textbf{72.27}    & \textbf{70.01}   & \textbf{75.99} & 68.59   & 61.80 & \textbf{69.73}   \\ \hline
\multicolumn{9}{|l|}{\textit{\textbf{Jointly Train on Both Source and Target Domains (Jointly Train)}}}                                                                                                    \\ \hline
\multirow{7}{*}{BERT-based} & \multicolumn{2}{c|}{w/o DAPT}                                                                     & 68.85    & 65.03   & 67.59 & 62.57   & 58.57 & 64.52  \\ \cline{2-9} 
                            & \multirow{3}{*}{\begin{tabular}[c]{@{}c@{}}DBpedia \\ Ontology-based\end{tabular}} & entity-level & 71.90     & 68.04   & 71.98 & 64.23   & 61.63 & 67.56  \\
                            &                                                                                    & task-level   & 71.31    & 67.75   & 71.17 & 63.24   & 60.83 & 66.86   \\
                            &                                                                                    & integrated   & \textbf{72.76}    & \textbf{68.28}   & \textbf{74.30}  & 65.18   & \textbf{63.07} & \textbf{68.72}  \\ \cline{2-9} 
                            & \multirow{3}{*}{\begin{tabular}[c]{@{}c@{}}NER \\ Model-based\end{tabular}}        & entity-level & 70.14    & 67.22   & 72.05 & 64.01   & 60.74 & 66.83   \\
                            &                                                                                    & task-level   & 71.10     & 66.89   & 70.92 & 63.59   & 59.36 & 66.37   \\
                            &                                                                                    & integrated   & 71.30     & 68.27   & 73,85 & \textbf{66.92}   & 62.15 & 67.16   \\ \hline
\end{tabular}
}
\caption{F1-scores (averaged over three runs) of proposed methods for the five domains in three settings. We use the span-level masking strategy for the DAPT.}
\label{supplementary_results}
\end{table*}

\section{Supplementary Experiments}
\subsection{Experimental Settings}
In experiments of the main paper, the entity list is created from the DBpedia Ontology for constructing the the entity-level corpus. And the domain-specialized entity list, which is needed to build the task-level corpus, is also based on the DBpedia Ontology. However, some low-resource domains might not be covered by DBpedia Ontology.
Therefore, to make the construction of the entity-level and task-level corpora more scalable, we utilize the NER model to categorize entities and substitute the DBpedia Ontology. 
Concretely, the process of generating entity-level and task-level corpus based on the NER model consists of three steps.
First, we leverage the NER training data in both source and target domains and follow the Pre-train then Fine-tune setting to build a NER model. Second, we use this trained NER model to recognize and categorize entities in the domain-related corpus. Third, we follow the same setting as that in the main paper to construct the entity-level and task-level corpora. In other words, we construct the entity-level corpus by extracting sentences having plentiful entities and create the task-level corpus by selecting sentences having domain-specialized entities. Finally, the integrated corpus is the combination of the entity-level and task-level corpora.
We make the size of the corpora constructed based on DBpedia Ontology and NER model comparable so as to have a fair comparison between these settings.

\subsection{Results \& Analysis}
The comparisons between the DBpedia Ontology and NER model are illustrated in Table~\ref{supplementary_results}. We can see that the DAPT using the NER model-based corpus achieves comparable results to using DBpedia Ontology-based corpus. The experimental results show that creating different levels of corpora does not rely on DBpedia Ontology.

% \bibliography{aaai2021}

\end{document}